# Back to the Drawing Board for Fair Representation Learning

**Angéline Pouget, Nikola Jovanović, Mark Vero, Robin Staab, Martin Vechev**
Department of Computer Science, ETH Zurich
{angeline.pouget,nikola.jovanovic}@inf.ethz.ch

## Abstract

The goal of Fair Representation Learning (FRL) is to mitigate biases in machine learning models by learning data representations that enable high accuracy on downstream tasks while minimizing discrimination based on sensitive attributes. The evaluation of FRL methods in many recent works primarily focuses on the tradeoff between downstream fairness and accuracy with respect to a single task that was used to approximate the utility of representations during training (*proxy task*). This incentivizes retaining only features relevant to the proxy task while discarding all other information. In extreme cases, this can cause the learned representations to collapse to a trivial, binary value, rendering them unusable in transfer settings. In this work, we argue that this approach is fundamentally mismatched with the original motivation of FRL, which arises from settings with many downstream tasks unknown at training time (*transfer tasks*). To remedy this, we propose to refocus the evaluation protocol of FRL methods primarily around the performance on transfer tasks. A key challenge when conducting such an evaluation is the lack of adequate benchmarks. We address this by formulating four criteria that a suitable evaluation procedure should fulfill. Based on these, we propose TransFair, a benchmark that satisfies these criteria, consisting of novel variations of popular FRL datasets with carefully calibrated transfer tasks. In this setting, we reevaluate state-of-the-art FRL methods, observing that they often overfit to the proxy task, which causes them to underperform on certain transfer tasks. We further highlight the importance of task-agnostic learning signals for FRL methods, as they can lead to more transferrable representations.

## 1 Introduction

The increased use of machine learning (ML) systems in critical decision-making processes has raised concerns about the fairness of these systems [21]. ML models have been found to perpetuate and exacerbate the biases present in the training data, inheriting the prejudices of prior decision-makers and disproportionately affecting certain demographic groups [2, 8, 12]. This is especially concerning in high-stakes environments such as criminal justice, healthcare, and hiring.

**Fair Representation Learning**  Companies operating in these environments generally collect large amounts of sensitive data, e.g., during the hiring process, when conducting user studies, or for clinical trials. Often, several teams working on different applications within the same company are interested in leveraging this data to improve their specific processes and products. In order to enable flexible usage of this data across the respective teams, it has to be ensured a priori that decisions derived from it will not be discriminatory against certain demographic groups. This poses a key challenge for algorithmic fairness. Fair Representation Learning (FRL) addresses this challenge by learning a data representation that removes indicators of protected group membership while retaining the information related to other features [28]. These representations can then safely be used by different teams to train models for various downstream applications. Crucially, such representations can be employed in practice if they (i) succeed at reducing the discriminatory impacts of the underlying data, and (ii) enable a sufficient level of utility across any downstream task.

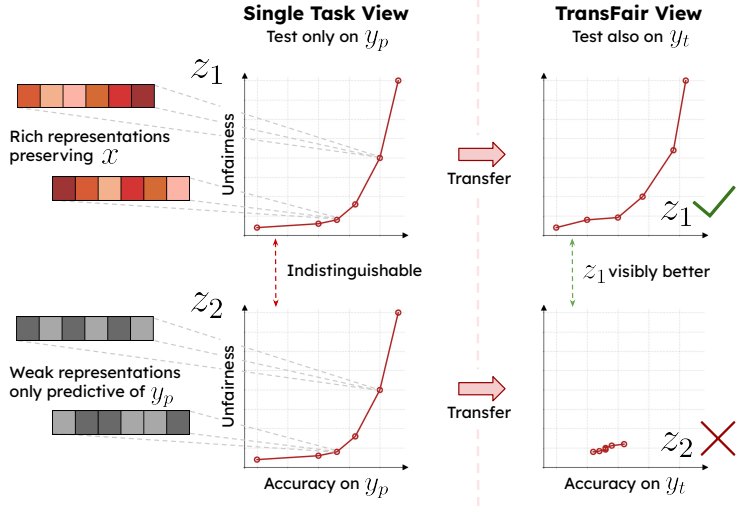

Figure 1: Most FRL works conduct their main evaluation by measuring the predictive performance of their learned representations on a proxy task $y_p$ that has also often been used to train the representations $z_i$. However, such an approach does not provide any insight into the downstream performance of the representations on other tasks. In fact, two representations $z_1$ and $z_2$, retaining vastly different amounts of information from the original data $x$, could still be indistinguishable under this evaluation method by being equally predictive of $y_p$. We introduce the TransFair benchmark to facilitate transfer evaluations on other tasks $y_t$, allowing one to successfully identify richer representations.

**Fundamental flaws in FRL evaluation** We argue that most current FRL methods presented in the literature fail to put sufficient focus on the aspect of universal utility in the learned representations (condition (ii) above). A majority of these methods evaluates the utility of the learned representations only on a single *proxy task*, often the task which was already available and used during training of the FRL method. This evaluation paradigm is fundamentally at odds with the original motivation of FRL. In particular, as the evaluation only involves a single proxy task that is known during training, there is no incentive for the representations to retain any additional information beyond what is necessary for this specific task. In such a setting, as illustrated in Fig. 1, two methods that are equally predictive of the proxy task will achieve the same performance, even if they retain vastly different amounts of information from the original data. As such, it becomes impossible for practitioners to identify transferrable representations that they can safely distribute to different teams.

**This work** In this work we address this gap, aiming to reconcile FRL evaluation with its original motivation. To achieve this, we propose an evaluation paradigm that favors methods with consistent performance across a systematically constructed set of downstream tasks unseen at training time (*transfer tasks*). This replaces the current evaluation protocol and enables practitioners to select representations that will be well-suited, even and especially for use-cases that cannot be anticipated at training time. While some prior studies have explored the need to evaluate methods on transfer tasks [19, 1, 13], such evaluations are often insufficient. This insufficiency arises because (i) these evaluations are rarely the primary focus, leading to the aforementioned pitfalls, and (ii) they commonly involve only a single transfer task per dataset, whose relationship to the proxy task is ambiguous, thereby hindering the extrapolation of the conclusions to other possible transfer tasks. To address these limitations, we formulate four criteria that a transfer benchmark facilitating such evaluations should fulfill. In particular, we emphasize the importance of equal treatment of all transfer tasks during evaluation. Based on these criteria, we propose the TransFair benchmark. In constructing TransFair, we extend and modify two popular fairness datasets, equipping them with four carefully calibrated transfer tasks each.

**Back to the drawing board**  Using the TransFair benchmark, we reevaluate prominent FRL methods and assess the utility of the learned representations across transfer tasks. While we find that all methods perform reasonably well on tasks highly correlated with the proxy task, their performance consistently degrades on weakly correlated or uncorrelated tasks. Notably, some supposedly fair representations lead to even higher unfairness than classifiers trained without any fairness considerations. This highlights the potentially harmful consequences of evaluating FRL methods on a single proxy task, as the resulting fairness may be greatly overestimated. At the same time, we observe that FRL methods that rely more on task-agnostic or unsupervised loss components than on the proxy task achieve favorable transferability, often outperforming methods that are considered state-of-the-art under the current evaluation protocol.

**Main contributions**  We summarize our key contributions:

- We point out the fundamental disconnect between the original motivation of FRL and the currently applied evaluation paradigms (§3).
- We propose four criteria that are essential for the development of effective FRL evaluation datasets and metrics, and introduce the TransFair benchmark that fulfills these criteria, enabling a more holistic evaluation of FRL methods (§4).
- We reevaluate existing FRL methods on TransFair, showing that the methods that rely solely on a proxy task fail to transfer to other weakly correlated or uncorrelated tasks (§5.1). We further demonstrate that methods with task-agnostic objectives on average exhibit more favorable transfer performance, and may be ultimately more suitable for real-world applications (§5.2).

## 2  Background: The Need for Fair Representation Learning

In this section, we introduce Fair Representation Learning (FRL) starting from its original motivation and intended form, and set up the notation used throughout the rest of the paper.

**Running example**  In line with popular work in FRL [19, 20, 28], we set up the following guiding example to serve as the main practical motivation for pre-processing sensitive data. A certain company collects large amounts of data about its users (e.g., personal information, user preferences, marketing and sales statistics). The company *leadership* wants to leverage this data to improve their products and user experience by sharing it with *teams* across the company (e.g., engineers, sales, marketing). These teams may then use this data for various predictive tasks. Crucially, as the data is sensitive, the leadership must ensure that any decisions derived from this data, automated or not, align with company policies and do not discriminate against any demographic group.

**The need for FRL**  One way to achieve this is via an *in-processing* approach, where each team trains their own machine learning model on raw data taking into account fairness. In this setup, the leadership would need to audit each model individually to ensure that they all adhere to the fairness criteria, which can become prohibitively complex and expensive. A more suitable and efficient solution comes in the form of *Fair Representation Learning (FRL)*. Here, the leadership (in the role of a *data regulator*) first defines the fairness criteria. Then, a designated team (*data producer*) is tasked with pre-processing the data into a debiased representation that can be safely shared with other teams (*data consumers*) for downstream usage. In this setup, there is no need to audit the downstream models individually, as their fairness can be estimated by analyzing the provided representations. In fact, for some methods [13], the data producer can even provide a rigorous guarantee on the maximum unfairness of any downstream model trained on the debiased representations. Thus, FRL conveniently shifts the responsibility of producing fair ML algorithms from a big set of potentially untrusted data consumers to a sole trusted data producer.

**Notation**  In this work, we focus on FRL for group fairness. Let $(\boldsymbol{x}, s) \in \mathbb{R}^d \times \mathcal{S}$ denote the original data tuples, where $s$ represents sensitive group membership of a user, and $\boldsymbol{x}$ a vector of features. The data producer trains a data encoder $f \colon \mathbb{R}^d \to \mathbb{R}^{d'}$, that maps each original data point into a representation $\boldsymbol{z} = f(\boldsymbol{x})$, aiming to remove the influence of the sensitive attribute. Each data consumer $i$ uses these representations to solve a different classification task, defined by a set of labels $y^{(i)} \in \mathcal{Y}^{(i)}$. The goal of each data consumer is to build a classifier $g^{(i)}$ that predicts $y^{(i)}$ from $z$, with no special considerations taken for fairness. We refer to any task not available during the training of $f$ as a *transfer task*.

**Maximizing fairness** This setup implies two competing goals that the data producer tries to achieve when training the encoder $f$. The first goal is to maximize fairness w.r.t. $s$, as measured by the fairness metric given by the data regulator (company leadership). Common choices include equal opportunity, equalized odds [12], or demographic parity [8]. In this work, we consider demographic parity (DP), precisely the DP-distance, which measures the maximum difference in expected outcomes between sensitive groups:

$$\Delta_{\text{DP}} \coloneqq \max_{(s_i, s_j) \in \mathcal{S} \times \mathcal{S}} |\mathbb{E}[g(\boldsymbol{x})|s_i] - \mathbb{E}[g(\boldsymbol{x})|s_j]|. \quad (1)$$

**Ensuring utility** The second goal is ensuring *utility*—the resulting representations $z$ should still be useful for the wide range of downstream tasks $y^{(i)}$ of the data consumers. This implies that the utility should ideally be defined in a task-agnostic way, i.e., in terms of mutual information between $x$ and $z$, which can be e.g., estimated as the success of a certain model at recovering $x$ from $z$. In exceptional cases, a data consumer can be involved in the FRL process. In this case, the utility can additionally be measured via the accuracy of a certain classifier on a *proxy task* $y_p$, i.e., the task of interest to this data consumer. The utility of that classifier is often judged w.r.t. the *unfair baseline*, another classifier trained directly on the original data samples $x$ without fairness considerations.

**The FRL objective** On a high-level, the most general FRL method represents the above goals of fairness and utility as three loss components: a fairness-aware loss $\mathcal{L}_f$, a task-agnostic loss $\mathcal{L}_r$, and a proxy task loss $\mathcal{L}_{y_p}$. While particular instantiations differ, these losses are commonly combined into a single training objective as follows:

$$\mathcal{L} = \lambda_f \mathcal{L}_f + \lambda_r \mathcal{L}_r + \lambda_{y_p} \mathcal{L}_{y_p}, \quad (2)$$

where $\lambda_f, \lambda_{y_p}, \lambda_r$ are hyperparameters that control the fairness-utility tradeoff. Many prominent FRL methods set $\lambda_r = 0$, focusing solely on the proxy task and ignoring the transferability of the learned representations. In the next section we elaborate on this choice, and make the case that it is flawed.

## 3 Current FRL Evaluation is Mismatched with its Motivation

Having established FRL from first principles, we now analyze the evaluation procedures used in prior FRL work from this perspective.

**The role of proxy tasks in current evaluations** As established in §2, a proxy task $y_p$ may be useful to quantify the utility of the learned representations during encoder training. Our key observation is that most prior work in the field goes beyond the assumption that a proxy task $y_p$ is available. In particular, the same proxy task is used as a primary signal when evaluating the fairness-utility tradeoff, and serves as the basis for comparing different FRL methods. This is often the only mode of evaluation [11, 18, 23], and only occasionally followed by a brief, separate evaluation of representations on other unseen tasks [28, 1, 13]. Motivated by our guiding example, which requires representations that achieve favorable fairness-utility tradeoffs across a wide range of (unseen) downstream tasks, we claim that this mode of evaluation is fundamentally mismatched with the motivation of FRL and can result in misleading conclusions, as we depict in Fig. 1. Instead, we argue that results on proxy tasks have to be presented alongside the results produced by the *same representations* on unseen tasks $y^{(i)}$.

**Focus on proxy tasks leads to in-processing** In support of this, we provide the following argument. To present results on a proxy task $y_p$ of some particular representation $z$, one trains a classifier $g$ to predict $y_p$ from $z$, and reports its accuracy and fairness w.r.t. the sensitive attribute $s$. In the absence of evaluation on unseen tasks, each such $z$ can be replaced by a binary representation that is equivalent to the prediction $g(z)$. This representation maintains both the accuracy and fairness of $z$ and would thus not impact the reported results. Moreover, $g(z)$ is a more favorable solution to the problem at hand, as the space of possible classifiers on these trivial representations is much smaller (4 in total), meaning that the data producer can accurately estimate the risk of unfairness in downstream classifiers. In fact, the most desirable representations in this setup are exactly those that best solve the proxy task (see Menon and Williamson [22] for a more thorough analysis), reducing FRL effectively to an *in-processing* problem. Such a solution strays far from the motivation of FRL and offers no insight regarding the performance in a real-world FRL scenario, as introduced in §2.

**The data gap** We believe that one key reason for this mismatch is the lack of suitable benchmarks for evaluating FRL methods in a way that reflects the original motivation of this setting, i.e., prioritizing transferability. The lack of benchmarks that meet the needs of current fairness research was already pointed out in recent works [6]—evaluation tools are even more scarce when it comes to specifically focusing on transfer capabilities of FRL methods. In §4, we aim to bridge this gap by formulating four criteria for a suitable FRL evaluation and by proposing *TransFair*, a principled transferability-aware FRL benchmark, comprised of two transfer-task extended datasets that fulfill our criteria. In §5, we evaluate prior work on TransFair and demonstrate the pitfalls of overreliance on proxy tasks.

## 4 Bridging the Evaluation Data Gap: TransFair

In this section, we address the lack of adequate transfer benchmarks for FRL evaluation by introducing the *TransFair* benchmark, a collection of FRL datasets with carefully calibrated transfer tasks and a corresponding suggested evaluation protocol. First, we define four detailed criteria a suitable transfer benchmark has to fulfill. Based on these criteria, we then propose transfer-extended versions of two popular tabular datasets in the fairness literature, presenting four transfer tasks for each of them in addition to their original proxy task. Finally, we describe the intended evaluation protocol for FRL methods on TransFair.

**Criteria For Transfer Datasets** To address the limitations of existing benchmarks and guide the curation of new FRL evaluation datasets, we propose the following criteria:

*C1—Dataset Size and Task Count:* The dataset should contain a sufficient number of samples across train and test splits to enable reliable training and testing of state-of-the-art machine learning models. Additionally, we require transfer evaluation datasets comprising a transfer benchmark to contain at least two tasks; a proxy task $y_p$ and one or more transfer tasks largely uncorrelated with $y_p$.

*C2—Sensitive Attribute:* The dataset should specify a canonical binary or categorical sensitive attribute such as gender, age, or race. This attribute should also be useful when solving each target task. If the sensitive attribute is uncorrelated with a task, discriminatory effects are likely not present even in fairness-unaware models. However, if the sensitive attribute is too strongly correlated with the target, achieving fairness might prove too difficult at any meaningful level of accuracy. Therefore, we propose a desired reasonable dependence on the sensitive attribute for the unfair baseline $u_t(x)$ on a given transfer task $y_t$ measured by a demographic parity distance between $0.05$ and $0.5$.

*C3—Correlations Between Tasks:* A wide range of FRL methods rely on a proxy task $y_p$ to obtain a training signal on utility. A fundamental limitation of current FRL evaluation is that the method is later also evaluated on $y_p$, which it may have overfit to. To approximate a worst-case view of the extent of this overfitting, we require a transfer benchmark to include at least one task that is approximately uncorrelated with $y_p$, i.e., a task where representations solely aimed at predicting $y_p$ would be ineffective. Including further tasks that are correlated with $y_p$ to varying degrees can enable additional insight into the influence of $y_p$ on the final representations. We measure the correlation between two tasks using the simple matching coefficient (SMC) [25]. An SMC of 100% corresponds to perfect correlation, while 50% indicates that two tasks are independent.

*C4—Appropriate Task Difficulty:* Each task included in the dataset should be of appropriate difficulty. Otherwise, differences across FRL methods and classification models may be too small, making it difficult to draw any conclusions about the tested methods. Although it is hard to anticipate what "appropriate difficulty" means in terms of accuracy on a given task, during the construction of TransFair, we observed that tasks with unfair baseline accuracies of between $70\%$ and $90\%$ are sufficiently informative. Additionally, the accuracy achieved by the unfair baseline should be sufficiently distinct from the accuracy of a constant predictor defaulting to the majority class (*majority baseline*). Here, we suggest a threshold of $> 5\%$ difference.

Table 1: Overview of the ACS-Transfer and Heritage-Health-Transfer tasks w.r.t. the criteria laid out in §4. UB denotes the *unfair baseline* and MB the *majority baseline*.

|  | Task | C2: UB Fairness | C3: SMC with $y_p$ | C4: UB Accuracy | C4: MB Accuracy |
|---|---|---|---|---|---|
| ACS Transfer | $y_p$: PINCP 50K | 0.065 | 100.0% | 80.0% | 64.2% |
|  | $y^{(1)}$: PERNP | 0.066 | 86.3% | 84.2% | 77.9% |
|  | $y^{(2)}$: PINCP 30K | 0.055 | 81.0% | 80.1% | 54.7% |
|  | $y^{(3)}$: JWMNP | 0.066 | 59.3% | 72.7% | 59.5% |
|  | $y^{(4)}$: WKW | 0.054 | 54.5% | 82.1% | 73.0% |
| Heritage-Health Transfer | $y_p$: max_CharlsonIndex | 0.358 | 100.0% | 75.1% | 68.0% |
|  | $y^{(1)}$: METAB3 | 0.394 | 70.3% | 78.4% | 65.1% |
|  | $y^{(2)}$: NEUMENT | 0.242 | 63.3% | 81.4% | 71.4% |
|  | $y^{(3)}$: ARTHSPIN | 0.174 | 61.9% | 79.4% | 67.9% |
|  | $y^{(4)}$: MSC2a3 | 0.201 | 50.0% | 78.4% | 61.9% |

**Dataset Construction** We select the ACS [6, 7] and Health Heritage [14] datasets as the basis for TransFair, as they both contain a sufficiently large amount of entries (C1) and have a canonical sensitive attribute used in the fairness literature (C2). For each dataset, we first select a set of features $x$ on which all prediction tasks will be performed, together with the proxy task $y_p$. Then, we select a set of candidate tasks from the remaining columns of the dataset $\mathcal{Y} = \{y^{(i)}\}_{i=1}^{l}$. Here, we make sure to include candidate tasks such that they satisfy C3, i.e., exhibit varying correlation levels to $y_p$, and at least one of the tasks is approximately uncorrelated with $y_p$. Finally, we train state-of-the-art tabular classifiers on each task and filter out any $y^{(i)}$ that does not comply with the task difficulty criterion C4 or with the baseline fairness criterion C2. As a result of our selection process, we identify four diverse transfer tasks for each of the datasets (in addition to their original proxy task). In the following, we provide details of the datasets contained in our TransFair benchmark.

**ACS-Transfer** We build upon the California-2014 subset of the ACS (American Community Survey) [6] PUMS (Public Use Microdata Sample) dataset provided by the United States Census Bureau, which consists of 372,553 data points derived from US-wide census data. This dataset is universally adopted in the fairness literature, with the *sex* feature treated as the canonical sensitive attribute $s$. We then filter this data to include only samples corresponding to individuals aged older than 16 and younger than 90 with an annual income over 100$, working at least 1 hour per week on average over the previous 12 months with a survey weight of at least 1 (this weighting ensures representative estimates when using the ACS PUMS dataset to infer information about US demographics). This filtering step ensures that we only keep samples that are sensible in the context of our considered tasks and leaves us with 183,896 samples (complying with C1). To construct our transfer dataset, we define a set of features detailed in App. D. Once we apply our filtering procedure described above on the remaining columns, we obtain 4 adequate transfer tasks related to individuals' earnings and employment. In Table 1, we show the transfer task statistics on the test split under criteria C2, C3, and C4 for the proxy task and each transfer task.

**Heritage-Health-Transfer** The Heritage Health dataset [14] contains 218,415 health records of patients' hospital stays (fulfilling C1). Its columns consist of both personal information (e.g., sex or age) and health indicators (detailed information in App. D), where the canonical sensitive attribute in the fairness literature is the age of the patients thresholded at 60 (C2). The main task, and as such the proxy task in most FRL works on this dataset, is to predict if maximum Charlson Comorbidity Index score observed for a patient over a specified period is non-zero. While certain prior works [19, 1, 13] have already employed this dataset to evaluate representations in a transfer setting, we are the first to carefully validate each potential transfer label and identify those most likely to provide valuable insights. To this end, we collect the transfer tasks used by these works and apply our selection criteria, leaving us with four suitable transfer tasks. In Table 1, we show the transfer task statistics on the test set w.r.t. our task selection criteria.

**Evaluation Protocol** The representations generated by a given FRL method (with or without access to a label) should be evaluated on all transfer labels for that dataset. To account for cases where certain models perform well on some labels but fail to achieve appropriate tradeoffs on other labels, the results for all transfer labels should be reported jointly. In particular, we suggest plotting a fairness-accuracy Pareto curve for each transfer label, where different combinations of hyperparameters lead to a different fairness-accuracy tradeoff, and presenting this alongside the results on the proxy task. We illustrate this using TransFair in our experimental evaluation in §5.

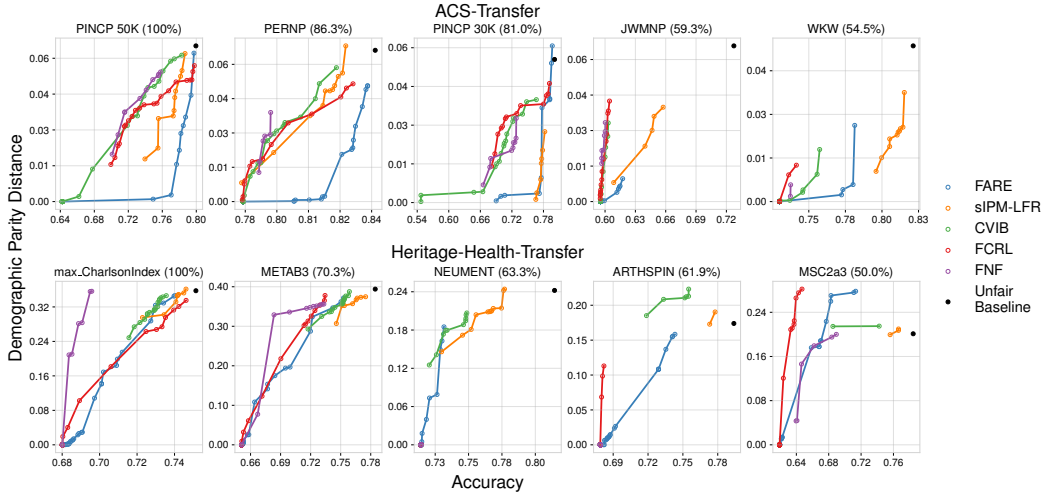

Figure 2: Accuracy-fairness Pareto fronts achieved by different FRL methods that rely solely on the proxy label $y_p$ to evaluate utility during training, on ACS-Transfer (*top*) and Heritage-Health-Transfer (*bottom*). Transfer tasks are sorted by decreasing correlation with $y_p$, shown as SMC in parentheses. The area shaded in red indicates representations with higher unfairness than the unfair baseline.

## 5 Experimental Evaluation on TransFair

In this section, we use our newly introduced TransFair benchmark to reevaluate state-of-the-art FRL methods, and investigate the impact of task-agnostic learning signals on their performance.

### 5.1 Reevaluation of FRL Methods

We reevaluate existing state-of-the-art supervised FRL methods on the new TransFair benchmark introduced in §4. In particular, we run each method on ACS-Transfer and Heritage-Health-Transfer with various hyperparameters to obtain different representations. For this, we follow the instructions provided in the respective writeups as well as Gupta et al. [11] to explore a dense parameter range (details in App. C). For each dataset, we use the respective proxy label $y_p$ as a learning signal during training. We then use a single-layer neural network with hidden layer size 50 trained on normalized representations as a downstream classifier. We train a separate classifier for all labels (proxy and transfer) and report the accuracy-fairness Pareto front for each. Following Gupta et al. [11], we train each classifier 5 times, reporting the average test set accuracy and the maximum DP distance. We include a range of popular methods: FARE [13], sIPM-LFR [16], CVIB [23], FCRL [11] and FNF [1]. Crucially, each of these methods relies solely on a single proxy label to evaluate the utility of the learned representations during training (for sIPM-LFR and CVIB, we set $\lambda_r = 0$). Learning these representations is fast for all considered methods, with training times being limited to at most 10 minutes. Models are trained either on a single NVIDIA GeForce RTX 2080 Ti GPU (sIPM-LFR, CVIB, FCRL and FNF) or on a single Intel(R) Xeon(R) Gold 6242 @ 2.80GHz CPU core (FARE).

**The impact of proxy task correlation** Our main results are presented in Fig. 2. As we would expect, all methods are most often able to sacrifice accuracy to improve fairness of the representations. While the exact values of the DP distance that are achievable for given accuracy vary across tasks (see C2 in §4), we shade the space of representations that are strictly worse than the unfair baseline in red. Our main observation is that the performance of each method varies significantly depending on the downstream task (see C3 in §4). On transfer tasks that are highly correlated with the proxy task used during training (such as PERNP on ACS-Transfer or METAB3 on Health-Heritage-Transfer), all methods perform reasonably well. However, when evaluating the representations on other tasks that are weakly or not at all correlated with the proxy label (such as WKW on ACS-Transfer or MSC2a3 on Heritage-Health-Transfer), performance starts to degrade significantly across all methods. This is especially true for the Health-Heritage-Transfer dataset, where learning to predict a weakly correlated label using the supposedly fair representations can lead to even higher demographic parity distances than the unfair baseline (red area in Fig. 2). This clearly indicates that learning and evaluating representations based on the single proxy label used during training is insufficient to ensure fairness across downstream tasks and can have potentially harmful consequences, as fairness is overestimated.

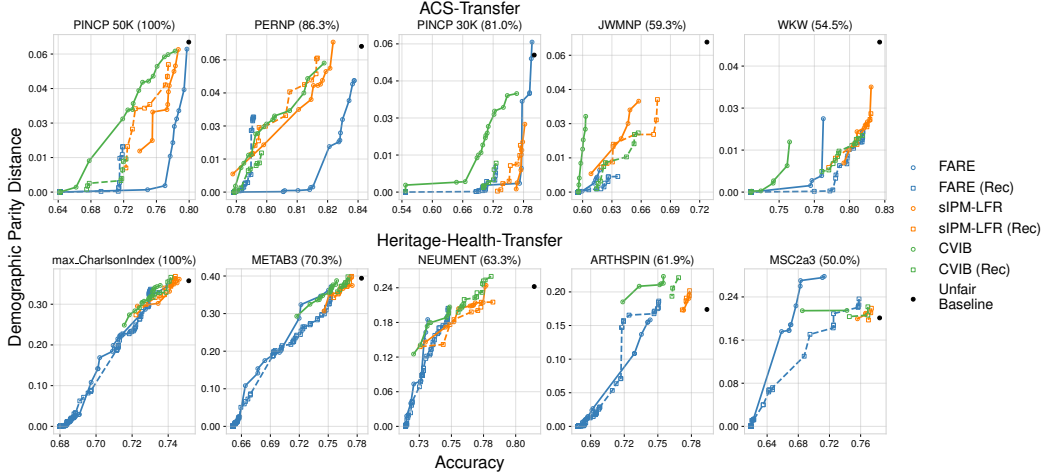

Figure 3: TransFair Pareto fronts achieved by FARE, sIPM-LFR and CVIB, with and without the reconstruction loss. **Top:** Results on ACS-Transfer, **Bottom:** Results on Heritage-Health-Transfer.

**Insufficiency of considering a single label** We additionally observe that it is not sufficient to compare FRL methods based on their performance on a single downstream task. While FARE seems to perform the best overall when representations are evaluated on the labels used during training, the representations learned by sIPM-LFR obtain comparable or slightly better results across other settings (e.g., JWMNP on ACS-Transfer). This suggests that the choice of FRL methods should be made based on downstream tasks that are expected to be encountered in practice and that this decision is significantly more complex and nuanced than assumed by current FRL evaluation paradigms.

### 5.2 Impact of Task-Agnostic Learning Signals

Based on these insights, a question that naturally arises is whether a task-agnostic learning signal ($\lambda_r > 0$) would help learn more generalizable representations. To this end, we compare sIPM-LFR and CVIB (with $\lambda_r = 0$) to *sIPM-LFR (Rec)* and *CVIB (Rec)* (with $\lambda_r > 0$ and $\lambda_y = 0$). We additionally introduce a reconstruction-based, task-agnostic version of FARE (*FARE (Rec)*) and compare it to the original FARE method (details in App. B).

**Double-edged impact of task-agnostic FRL** The results of this additional evaluation are shown in Fig. 3. We primarily observe that adding the reconstruction loss to FARE leads to a significant improvement in the performance of on those transfer tasks that are weakly correlated or uncorrelated to the proxy task. This suggests that the reconstruction loss can help to learn more task-agnostic representations that are more robust across a range of downstream tasks. However, the reconstruction loss also leads to a decrease in performance on the transfer tasks that are highly correlated with the proxy task. While we can see minor improvements in the performance of sIPM-LFR and CVIB on certain transfer tasks (see WKW on ACS-Transfer), the addition of a task-agnostic loss component seems to be insufficient to learn useful representations for others (see MSC2a3 on Heritage-Health-Transfer). It should be emphasized that the choice between a task-specific and task-agnostic loss is by no means a binary one and in many settings a combination of both might be beneficial, leading to a more robust and generalizable model.

## 6 Related Work

In this section, we provide a brief overview over relevant related work in the broader field of fair machine learning and a more detailed overview over fair representation learning methods.

**In- and Post-processing for Fair Machine Learning**  Orthogonal to learning fair representations, a long line of work focuses on approaches that directly modify the training of a prediction algorithm [10, 15, 27]. This is commonly achieved by adding fairness regularization to the overall training objective or directly enforcing harder constraints on the full optimization [27, 10]. Similarly, there exist works on post-processing approaches that, instead of intervening in the training of a classifier, modify the decision boundaries of an already trained model [5, 12, 3]. A key issue for both in- and post-processing approaches is that their effect is limited to the intervened model, requiring potentially expensive retraining and fairness evaluations for any new model and task.

**Fair Representation Learning**  To address this challenge, work in fair representation learning has gained significant interest in recent years [1, 11, 18, 20, 24]. Popular variational auto-encoder (VAE)-based approaches [11, 18] learn fair representations by directly minimizing the information encoded in them while maintaining their usefulness for a downstream task. This is extended by approaches such as [4] that regularize the VAE latent space to disentangle individual attributes. Several works [9, 19, 26, 17] instead propose learning representation via adversarial training, jointly learning an encoder alongside an adversary. Kim et al. [16] builds on this setting using integral probability metrics. Recent methods building on normalizing flows [1] and restricted encoders [13] further allow for strict guarantees on the maximum fairness violation of any downstream classifier. Notably, we are unaware of any work that primarily focuses on the evaluation procedure and datasets used for realistic FRL. With the exception of McNamara et al. [20], which describes a more complete picture of the FRL pipeline, most prior work mainly evaluates their proposed FRL algorithms on the proxy task used during training with no standardized evaluation protocol across writeups.

## 7 Limitations

While our work highlights the importance of reassessing FRL evaluation paradigms and evaluating methods on multiple downstream tasks, several limitations remain. First, we focus on a limited set of two FRL datasets which may not fully represent the diversity of tasks FRL could be applied to. In particular, an interesting avenue for future work would be the creation of a novel dataset from scratch, tailored specifically to FRL. Second, while we demonstrate a viable direction towards more transferrable FRL (i.e., through introduction of task-agnostic loss components), it remains to be seen if this is sufficient to match performance of task-specific FRL methods that are evaluated on the same task they were trained on. Third, we only consider a single fairness metric, demographic parity distance—our analysis can be directly extended to include other metrics such as equal opportunity or equality of odds. Finally, our evaluation of representations on downstream tasks is currently limited to single-layer neural networks. It would be interesting to investigate the performance of more complex models on the learned representations and to explore potential interactions between FRL methods and downstream model architectures.

## 8 Conclusion

This work highlights a fundamental mismatch between the current evaluation practices for FRL methods, which is generally based on a single proxy task used during both training and evaluation, and the foundational goals of FRL. We advocate for an evaluation protocol centered around performance on multiple transfer tasks, reflecting the real-world scenarios FRL is designed to handle. To this end, we introduce a set of criteria to guide future FRL dataset creation. Moreover, we introduce TransFair, a collection of FRL datasets with carefully calibrated transfer tasks meant to inform the development and evaluation of novel FRL methods going forward. We see our work and particularly the TransFair benchmark as an important step towards more realistic and robust evaluations of FRL algorithms.

## A More Results

In the following, we present an additional result that was omitted from the main paper for brevity.

**Effect of directly using a transfer label during training** There are certain labels such as JWMNP on ACS-Transfer or MSC2a3 on Heritage-Health-Transfer for which all FRL methods perform quite poorly. While this might suggest that these tasks are inherently more difficult, we show in Fig. 4 that this is not the case. When training FARE directly on the label in question, we obtain a reasonable fairness-utility tradeoff for all transfer tasks. In addition, this experiment also nicely visualizes the impact of correlation between the proxy and a transfer task with the difference between *FARE (Eval)* and *FARE (Proxy)* generally increasing for more weakly correlated labels.

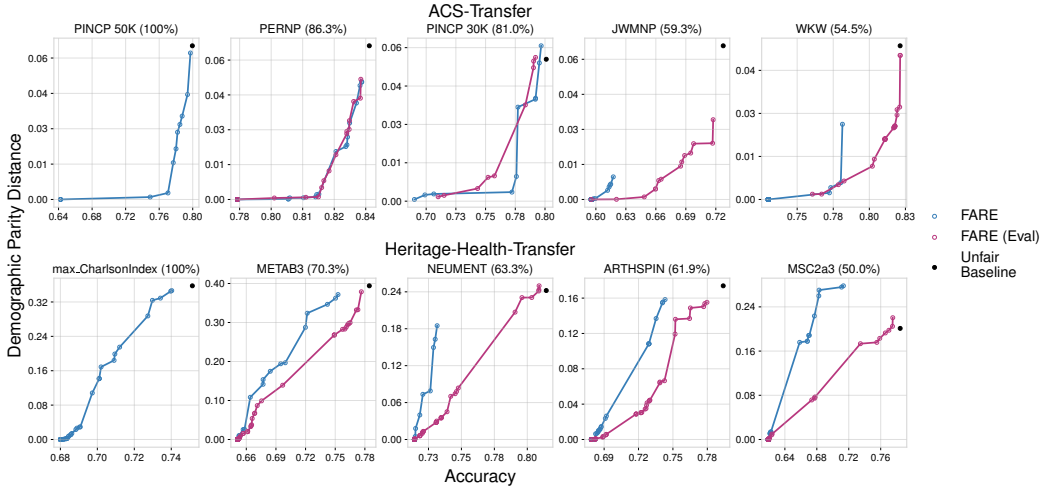

Figure 4: As can be seen, it is possible to obtain a reasonable fairness-utility tradeoff for all labels in principle. For the purpose of demonstrating this, we train *FARE (Eval)* directly using the transfer label in question during training and compare this to *FARE (Proxy)* that has been trained on the proxy task $y_p$.

## B FARE with Reconstruction Loss

As mentioned in §5, we introduce an alternative loss for FARE [13]. FARE is a state-of-the-art FRL method that is based on fair classification trees used as restricted encoders. During training, leaves are split with the aim of minimizing $FairGini(D) = (1 - \gamma)Gini_y(D) + \gamma(0.5 - Gini_s(D))$ with $\gamma$ being a hyperparameter that balances the importance of the fairness and utility components. While $Gini_y(D)$ is the standard Gini impurity for the labels, $Gini_s(D)$ is a measure of the impurity of the sensitive attribute. Once the tree is fully constructed, all data points in a given leaf are reduced to their median, which is then used as the representation of that leaf.

This setup naturally lends itself to augmentation with a reconstruction-based loss used to encourage splits that lead to similar data points being in the same leaf. We calculate this loss by comparing an original data point $x_i$ with the mean of all data points in leaf $j$, $\hat{x}_j = \sum_{i=1}^{n_j} x_i/n_j$. We define the loss as the mean squared error $\sum_{i=1}^{n_j} \|x_i - \hat{x}_j\|^2/n_j$. We alternatively also consider a loss based on the absolute distance to the median of all points in a given leaf and find that the results were similar (see Fig. 5 for a comparison).

## C Experimental Evaluation Details

As mentioned in §5, we consider a range of hyperparameters for each FRL method. For CVIB (both standard and task-agnostic), we explore $\lambda \in [0.01, 1]$ and $\beta \in [0.001, 0.1]$. For FCRL, we explore $\lambda = \beta \in [0.02, 2]$. For sIPM-LFR, we use $\lambda \in [0.0001, 1], \lambda_r = 0$ (standard) and $\lambda_r \in [0.0001, 1], \lambda = 0$ (task-agnostic) with $\lambda_f \in [0.0001, 100]$ for both. For original FARE, there are four hyperparameters: $\gamma$ (used for the criterion, where larger $\gamma$ puts more focus on fairness), $\bar{k}$ (upper bound for the number of leaves), $n_i$ (lower bound for the number of examples in a leaf), and $v$ (the ratio of the training set to be used as a validation set). We explore $\gamma \in [0, 1]$ and $k \in [2, 200]$ and set the other parameters to $n_i = 100$ and $v = 0.3$ due to limited impact on the resulting representation. For FARE (Rec), we additionally explore $k \in [200, 12800]$, $\lambda_f \in [0.1, 1]$, $\lambda_r \in [0.001, 1000]$ and set $\lambda_y = 0$. It should be noted that for $k > 200$, the computation of the upper bound is generally no longer possible and, hence, in this setting FARE does not offer any fairness guarantees.

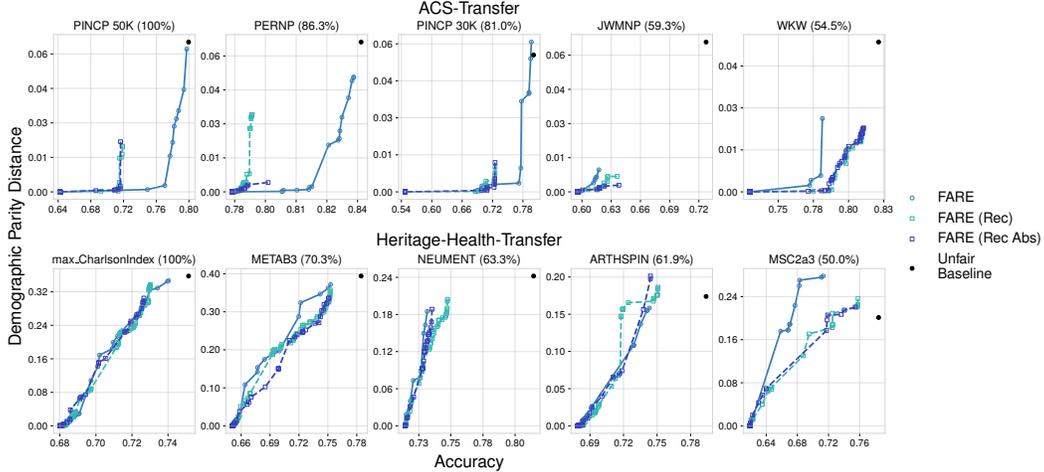

Figure 5: Comparison of FARE with the reconstruction loss based on the mean squared distance (FARE (Rec)) and the absolute distance (FARE (Rec Abs)) on all transfer tasks. The top row shows the results on ACS-Transfer, while the bottom row shows the results on Heritage-Health-Transfer.

## D  Details of the TransFair Datasets

This section provides additional information about ACS-Transfer and Heritage-Health-Transfer datasets introduced in §4. It includes a detailed description of the features and labels present in the datasets. For ACS-Transfer, the data itself is governed by the terms of use provided by the Census Bureau. For our modification, we use the [Folktables](https://github.com/socialfoundations/folktables/tree/main) that is released under the MIT license. Heritage Health was first released as part of the [Heritage Health Prize Contest](https://www.kaggle.com/competitions/hhp/overview).

Table 2: Overview of ACS-Transfer features and labels.

| Name | Description | Values |
| --- | --- | --- |
| Features $x$ | | |
| AGEP | Person's age | 0 to 99 |
| ANC | Ancestry or ethnic origin | 1 = Single, 2 = Multiple, 3 = Unclassified, 4 = Not reported |
| CIT | Citizenship status | 1 = Born in US, 2 = Born in US territory, 3 = Born abroad of US parents, 4 = Naturalized citizen, 5 = Not a citizen |
| COW | Class of worker | Various codes for private company, government, or self-employment |
| DEAR | Difficulty hearing | 1 = Yes, 2 = No |
| DEYE | Difficulty seeing | 1 = Yes, 2 = No |
| DIS | Disability status | 1 = Yes, 2 = No |
| DREM | Cognitive difficulty | 1 = Yes, 2 = No |
| ESP | Employment status of parents | Various codes indicating if both, one or no parent is in labor force |
| JWTR | Means of transportation to work | Various codes representing different modes of transportation |
| MAR | Marital status | 1 = Married, 2 = Widowed, 3 = Divorced, 4 = Separated, 5 = Never married |
| NATIVITY | Whether a person is native or foreign-born | 1 = Native, 2 = Foreign-born |
| RAC1P | Race | Various codes representing different races |
| RELP | Relationship to the head of household | Various codes representing different relationships |
| SCHL | Educational attainment | Various codes representing different education levels |
| SEX | Gender | 1 = Male, 2 = Female |
| WKHP | Usual hours worked per week in the past 12 months | bb = Not working, 0-98 = 1-98, 99 = 99 or more |
| PUMA | Public Use Microdata Area code | Codes representing geographic areas |
| POWPUMA | Place of work Public Use Microdata Area code | Codes representing geographic areas |
| Labels $y$ | | |
| PINCP 50K | Total person's income (past 12 months) above 50k | 0 = No, 1 = Yes |
| PERNP | Total person's earnings (past 12 months) above 70k | 0 = No, 1 = Yes |
| PINCP 30K | Total person's income (past 12 months) above 30k | 0 = No, 1 = Yes |
| JWMNP | Travel time to work above 20 minutes | 0 = No, 1 = Yes |
| WKW | Weeks worked in the past 12 months | 0 = Less than 50 weeks, 1 = 50 to 52 weeks |

Table 3: Overview of Health-Heritage-Transfer features and labels.

| Name | Description | Values |
|---|---|---|
| \multicolumn{3}{c}{Features $x$} | | |
| LabCount_total | Total number of lab tests conducted | Non-negative integer |
| LabCount_months | Number of months with lab tests conducted | Non-negative integer |
| DrugCount_total | Total number of drugs prescribed | Non-negative integer |
| DrugCount_months | Number of months with drugs prescribed | Non-negative integer |
| no_Claims | Number of claims made | Non-negative integer |
| no_Providers | Number of unique healthcare providers visited | Non-negative integer |
| no_Vendors | Number of unique vendors visited | Non-negative integer |
| no_PCPs | Number of unique primary care physicians visited | Non-negative integer |
| PayDelay_total | Total payment delay (in days) | Non-negative integer |
| PayDelay_max | Maximum payment delay (in days) | Non-negative integer |
| PayDelay_min | Minimum payment delay (in days) | Non-negative integer |
| Specialty=Anesthesiology | Indicates if specialty is Anesthesiology | 0 = No, 1 = Yes |
| Specialty=Diagnostic Imaging | Indicates if specialty is Diagnostic Imaging | 0 = No, 1 = Yes |
| Specialty=Emergency | Indicates if specialty is Emergency | 0 = No, 1 = Yes |
| Specialty=General Practice | Indicates if specialty is General Practice | 0 = No, 1 = Yes |
| Specialty=Internal | Indicates if specialty is Internal | 0 = No, 1 = Yes |
| Specialty=Laboratory | Indicates if specialty is Laboratory | 0 = No, 1 = Yes |
| Specialty=Obstetrics and Gynecology | Indicates if specialty is Obstetrics and Gynecology | 0 = No, 1 = Yes |
| Specialty=Other | Indicates if specialty is Other | 0 = No, 1 = Yes |
| Specialty=Pathology | Indicates if specialty is Pathology | 0 = No, 1 = Yes |
| Specialty=Pediatrics | Indicates if specialty is Pediatrics | 0 = No, 1 = Yes |
| Specialty=Rehabilitation | Indicates if specialty is Rehabilitation | 0 = No, 1 = Yes |
| Specialty=Specialty_? | Indicates if specialty is unknown | 0 = No, 1 = Yes |
| Specialty=Surgery | Indicates if specialty is Surgery | 0 = No, 1 = Yes |
| ProcedureGroup=ANES | Indicates if procedure group is ANES | 0 = No, 1 = Yes |
| ProcedureGroup=EM | Indicates if procedure group is EM | 0 = No, 1 = Yes |
| ProcedureGroup=MED | Indicates if procedure group is MED | 0 = No, 1 = Yes |
| ProcedureGroup=PL | Indicates if procedure group is PL | 0 = No, 1 = Yes |
| ProcedureGroup=ProcedureGroup_? | Indicates if procedure group is unknown | 0 = No, 1 = Yes |
| ProcedureGroup=RAD | Indicates if procedure group is RAD | 0 = No, 1 = Yes |
| ProcedureGroup=SAS | Indicates if procedure group is SAS | 0 = No, 1 = Yes |
| ProcedureGroup=SCS | Indicates if procedure group is SCS | 0 = No, 1 = Yes |
| ProcedureGroup=SDS | Indicates if procedure group is SDS | 0 = No, 1 = Yes |
| ProcedureGroup=SEOA | Indicates if procedure group is SEOA | 0 = No, 1 = Yes |
| ProcedureGroup=SGS | Indicates if procedure group is SGS | 0 = No, 1 = Yes |
| ProcedureGroup=SIS | Indicates if procedure group is SIS | 0 = No, 1 = Yes |
| ProcedureGroup=SMCD | Indicates if procedure group is SMCD | 0 = No, 1 = Yes |
| ProcedureGroup=SMS | Indicates if procedure group is SMS | 0 = No, 1 = Yes |
| ProcedureGroup=SNS | Indicates if procedure group is SNS | 0 = No, 1 = Yes |
| ProcedureGroup=SO | Indicates if procedure group is SO | 0 = No, 1 = Yes |
| ProcedureGroup=SRS | Indicates if procedure group is SRS | 0 = No, 1 = Yes |
| ProcedureGroup=SUS | Indicates if procedure group is SUS | 0 = No, 1 = Yes |
| PlaceSvc=Ambulance | Indicates if place service is Ambulance | 0 = No, 1 = Yes |
| PlaceSvc=Home | Indicates if place service is Home | 0 = No, 1 = Yes |
| PlaceSvc=Independent Lab | Indicates if place service is Independent Lab | 0 = No, 1 = Yes |
| PlaceSvc=Inpatient Hospital | Indicates if place service is Inpatient Hospital | 0 = No, 1 = Yes |
| PlaceSvc=Office | Indicates if place service is Office | 0 = No, 1 = Yes |
| PlaceSvc=Other | Indicates if place service is Other | 0 = No, 1 = Yes |
| PlaceSvc=Outpatient Hospital | Indicates if place service is Outpatient Hospital | 0 = No, 1 = Yes |
| PlaceSvc=PlaceSvc_? | Indicates if place service is unknown | 0 = No, 1 = Yes |
| PlaceSvc=Urgent Care | Indicates if place service is Urgent Care | 0 = No, 1 = Yes |
| Sex | Gender | 1 = Male, 2 = Female |
| \multicolumn{3}{c}{Labels $y$} | | |
| max_CharlsonIndex | Maximum Charlson Comorbidity Index | Non-negative integer |
| MSC2a3 | Indicates presence of MSC2a3 condition | 0 = No, 1 = Yes |
| METAB3 | Indicates presence of METAB3 condition | 0 = No, 1 = Yes |
| ARTHSPIN | Indicates presence of ARTHSPIN condition | 0 = No, 1 = Yes |
| NEUMENT | Indicates presence of NEUMENT condition | 0 = No, 1 = Yes |